# Conditional GraphGANFed: Optimizing Graph-Structured Molecule Generation in Federated Generative Adversarial Networks

Daniel Manu, and Abee Alazzwi

***Abstract*—Generative adversarial networks (GANs) have garnered considerable attention in molecular discovery for their ability to generate novel and high-quality molecules. To efficiently train a GAN model while preserving data privacy, GraphGANFed has been proposed to incorporate federated learning and graph convolutional networks into GAN. Yet, GraphGANFed cannot produce synthetic molecules that only optimize a user-defined metric(s) to facilitate the new drug discovery process. To address this issue, we introduce a novel extension to GraphGANFed, namely conditional GraphGANFed (cGraphGANFed), by incorporating the critic network to assess generated molecules using user-defined metric(s). The evaluation results from both the critic network and discriminator are integrated into the loss function of the generator, guiding it to generate novel molecules that maintain similar chemical properties to real ones while optimizing user-defined metrics. Extensive simulations are conducted in two scenarios. First, cGraphGANFed endeavors to optimize all seven commonly used metrics, and the results show that cGraphGANFed significantly outperforms GraphGANFed in Validity and LogP, with a slight advantage in QED, across different settings. Second, cGraphGANFed focuses solely on optimizing QED, and the results show that the synthetic molecules produced by cGraphGANFed can achieve more than 10% improvement in QED than GraphGANFed. Also, the results demonstrate cGraphGANFed has enhanced resilience against mode collapses and performance reduction caused by non-IID data.**

***Index Terms*—Generative adversarial networks, Graph convolutional networks, Federated learning, Drug discovery**

## I. Introduction

In the field of drug discovery and design, the search for novel molecules with optimized chemical properties remains a formidable challenge. Traditional methods face hurdles in efficiently exploring the vast chemical space. Moreover, the discrete nature of chemical space means even minor alterations in molecular structure can yield significant changes in desired properties [1, 2]. Current approaches for exploring and discovering molecules involve high-throughput virtual screening or discrete local search methods like genetic algorithms [3, 4]. High-throughput screening involves generating an extensive compound library, followed by successive filtration using increasingly sophisticated computational techniques to assess if a given species exhibits the desired properties. While this method has achieved notable successes in pharmaceutical applications [5, 6], it suffers from substantial limitations. Specifically, its exploration of the molecular space is constrained by the combinations initially present in fragment libraries.

Enhancing the efficiency of generating new molecules tailored to specific application-driven objectives could significantly accelerate progress in developing innovative drugs. In such applications, guiding the generative model towards some desirable properties while preserving resemblance to the original distribution proves highly effective [1, 7, 8]. Recent advancements have propelled the utilization of deep learning models in molecular generation, yet the challenge persists in producing valid and innovative molecular structures that effectively optimize various physical, chemical, and biological properties due to the inherent complexity and non-differentiability of these property objectives [9, 10]. Deep generative models have demonstrated versatility across a spectrum of domains including natural language processing [11, 12] and biomedicine [13–16]. In the field of drug discovery, deep generative models learn probability distributions over extensive molecular sets and facilitate the generation of novel molecules through sampling from these distributions.

Generative adversarial networks (GANs), a prominent deep generative model, have gained significant traction in molecular design due to their capacity to produce highly realistic content [17]. GAN comprises a generator and discriminator that compete against each other during training. The generator strives to produce synthetic samples closely resembling real ones, thereby fooling the discriminator. Meanwhile, the discriminator tries to distinguish between synthetic and real samples. The objective-reinforced GAN (ORGAN) framework [18] pioneered the use of GANs for molecular generation, applying the SMILES representation as input data. Its improved version, ORGANIC [10], is designed specifically for the direct inverse design of molecules. Subsequently, the molecular GAN (molGAN) framework [9] emerged, focusing on molecular generation, where molecules are represented by molecular graph representations as opposed to SMILES. However, these approaches are limited in their ability to generate molecules tailored specifically to optimize certain desired properties. To address this limitation, a combination of RL and GANs has been proposed. This fusion aims to create molecular structures that precisely optimize predefined properties. The SeqGAN architecture [12] exemplifies this fusion, extending the GAN framework with an RL-based generator. It updates the generator's parameters using policy

D. Manu, A. Alazzwi, and X. Sun are with the SECNet Lab., Department of Electrical and Computer Engineering, University of New Mexico, Albuquerque, NM 87131, USA. E-mail: {dmanu,abeer88,sunxiang}@unm.edu.

This work was supported by the National Science Foundation under Award under grant no. CNS-2323050 and CNS-2148178, where CNS-2148178 is supported in part by funds from federal agency and industry partners as specified in the Resilient & Intelligent NextG Systems (RINGS) program.

gradients and Monte Carlo search [19], guided by expected rewards from the discriminator. The REINFORCE algorithm [20] was used to estimate the action-value function. Putin *et al.* [21] showed the effectiveness of the proposed Adversarial Threshold Neural Computer (ATNC) framework integrating GANs and RL. By leveraging adversarial thresholds, ATNC successfully addressed the negative reward problem in ORGANIC while generating target molecules.

Cutting-edge molecular generation frameworks integrating RL optimization have shown promising results in generating molecules with optimized metrics efficiently [9, 10, 18]. However, training these frameworks demands significant and varied molecular datasets. Acquiring such comprehensive datasets is hindered by stringent data regulations and privacy concerns prevalent in the pharmaceutical industry [22, 23]. Secure multiparty computation (MPC) [24] offers a contemporary solution for maintaining privacy while facilitating collaboration among pharmaceutical companies. MPC enables multiple parties in the pharmaceutical sector to share their encrypted molecular datasets with a trusted third party. Nonetheless, challenges such as computational complexity, communication overhead, and data privacy concerns persist. Federated Learning (FL) emerges as a promising approach for training deep learning models in a distributed fashion while preserving data privacy. By leveraging FL, clients in terms of pharmaceutical companies can train their models using their respective datasets locally. They then upload only their model parameters to a centralized FL server, eliminating the need for sharing their datasets. Therefore, the integration of FL into the GAN training process can enhance scalability by utilizing diverse datasets and computing power from multiple clients, thus improving GAN performance without compromising data privacy [22, 23]. While the integration of FL, GAN, and RL shows promise, there's currently a lack of research in this area. Additionally, the integration is complex because it involves multiple deep learning models with different loss functions that interact during the training process. For example, a delicate balance is required between the performance of the generator and discriminator in a GAN to achieve stable and convergent training. Any imbalance, where one model overpowers the other, risks destabilizing or even diverging the training process.

In this paper, we propose a novel framework, namely conditional GraphGANFed (cGraphGANFed), that applies FL to train the framework that incorporates the critic network and graph convolutional network into GAN, using the real molecules represented as graphs. cGraphGANFed can generate synthetic molecules that not only preserve similar chemical properties as the real ones but also optimize user-defined metrics to facilitate the new drug discovery process without compromising data privacy. The main contributions of the paper are summarized as follows:

- We propose a novel cGraphGANFed framework, which is developed based on our previously proposed framework, GraphGANFed [14]. cGraphGANFed incorporates the critic network in RL to assess generated molecules using user-defined metric(s). The evaluation results from both the critic network and discriminator are incorporated into the loss function of the generator, guiding it to generate novel molecules that maintain similar chemical properties to real ones while optimizing user-defined metrics.
- To preserve data privacy during the model training, cGraphGANFed also applies FL to avoid the sharing of molecule samples among different clients in terms of pharmaceutical companies.
- Extensive simulations are conducted to demonstrate the performance of cGraphGANFed under the following two scenarios, (a) optimization of a specific molecular metric, and (b) optimization of all molecular metrics simultaneously. The results show that cGraphGANFed can (a) produce molecules with significantly higher values of the specified molecular metric in comparison to other metrics, or (b) generate the molecules that exhibit high values across the majority of molecular metrics, which have dependent relationships (e.g., QED, Validity, and LogP).

The rest of the paper is organized as follows. Section II presents a brief background of GANs and related works regarding the integration of RL into molecular generation. Section III introduces the GAN-based molecular generation problem formulation, cGraphGANFed architecture, and optimization the molecular generation under federated settings. Section IV discusses the simulation setups and results. Finally, Section V concludes the paper.

## II. Background and Related Works

In this section, we provide a concise introduction to GANs and conduct a survey of existing frameworks that employ RL to optimize molecular generation.

### A. Background of GANs

GANs aim to minimize the discrepancy between the synthetic data distribution $(P_s)$ and the real data distribution $(P_r)$. A GAN framework comprises a generator and discriminator, both engaging in a competitive training process. The generator utilizes random noise vector $\boldsymbol{z}$ to produce diverse synthetic samples that resemble real samples, while the discriminator learns to distinguish between real and synthetic samples. The generator's objective is to minimize the loss function $log(1-D(\tilde{x}))$, where $\tilde{x}$ represents a synthetic sample output by the generator, i.e. $\tilde{x} = G(\boldsymbol{z})$, and $D(\tilde{x})$ is the discriminator's output, indicating the estimated score/probability that synthetic sample $\tilde{x}$ is real. Conversely, the discriminator aims to maximize its accuracy in discriminating between real and synthetic samples. The standard GAN loss function, also referred to as the min-max loss, is defined as [25, 26]

$$\min_{\theta} \max_{\varphi} \mathbb{E}_{x\sim P_r}[logD_{\varphi}(x)]+\mathbb{E}_{\tilde{x}\sim P_s}[log(1-D_{\varphi}(G_{\theta}(z))]. \quad (1)$$

Here, $G_{\theta}(\cdot)$ and $D_{\varphi}(\cdot)$ are the generator and discriminator models, respectively, parameterized by $\theta$ and $\varphi$. $\mathbb{E}_{x\sim P_r}[logD_{\varphi}(x)]$ is the expected value computed over all real samples, reflecting the probabilities that the discriminator accurately classifies the real samples. Similarly, $\mathbb{E}_{\tilde{x}\sim Ps}[log(1 - D_{\varphi}(G_{\theta}(z))]$ is the expected value computed over all generated samples from the generator, reflecting the probabilities that the discriminator accurately classifies the synthetic samples.

### B. RL-based Molecular Generative Frameworks

Traditional GANs face a significant limitation in generating synthetic samples tailored to optimize some user-defined metrics. To address this challenge, several approaches have proposed integrating RL with GANs, particularly in molecular generation applications. For instance, [18] introduces the ORGAN framework, which integrates RL into GANs for sequence molecular generation. ORGAN employs a policy gradient-based approach within the GAN framework to optimize molecular generation towards specific objectives. ORGAN focuses on the inverse design of molecules incorporating rewards from the discriminator to guide the generator. Using a Monte Carlo tree search and policy gradients, it updates the generator to maximize the expected reward. ORGAN addressed the challenge of discrete molecular spaces by leveraging RL policy, enabling the generation of novel molecules while addressing the negative reward problem. You *et al.* [1] proposed a novel framework using graph convolutional networks (GCNs) to generate molecular graphs towards specific objectives. They used a policy network to learn a probability distribution over actions for graph generation, enabling goal-directed molecule creation. Using an RL framework, the model leverages rewards from a discriminator to guide the generation process. The approach combines graph convolutions with policy gradients, enhancing molecule generation by effectively navigating through the chemical space. The results demonstrated the capability to optimize molecular graphs towards desired properties while overcoming challenges associated with discrete molecular representations. Pereira *et al.* [27] presented a method leveraging RL to enhance diversity in generating targeted molecules. It incorporates a diversity-promoting term in the reward function to encourage the exploration of the chemical space. Through policy gradients, the model learned to balance between maximizing target property and promoting diversity in generated molecules. This approach solved issues of generating only similar molecules with high property values. By integrating diversity improvement within RL, the model achieved improved exploration of molecular space while optimizing towards specific properties.

## III. Conditional GraphGANFed: Optimizing Graph-Structured Molecule Generation in Federated Generative Adversarial Networks (cGraphGANFed)

In this section, we will provide details of the user-defined metric optimization, explain the detailed design of the proposed cGraphGANFed architecture and introduce the metrics to evaluate the synthetic molecules.

### A. User-defined Metric Optimization in Molecule Discovery

In GANs, the generator is trained to map a prior distribution to the real data distribution, resulting in synthetic samples that closely resemble those from the real data distribution. However, in de novo drug design, the objective extends beyond merely generating chemically valid molecules to ensuring that these molecules possess specific useful properties, such as synthesizability. In a typical Wasserstein GAN (WGAN), the generator and discriminator loss functions are defined as

$$\mathcal{L}^{gen} = -\mathbb{E}[\log D_{\varphi}(\tilde{x})] \quad (2)$$

$$\mathcal{L}^{dis} = \underbrace{-\mathbb{E}[D_{\varphi}(x)] + \mathbb{E}[D_{\varphi}(\tilde{x})]}_{\text{Discriminator loss}} + \underbrace{\gamma\mathbb{E}[(\|\nabla_{\hat{x}} D_{\varphi}(\hat{x})\|_2 - 1)^2]}_{\text{Gradient penalty}}, \quad (3)$$

where $x$, $\tilde{x}$, and $\hat{x}$ represent a real, synthetic, and random molecule that is sampled from a set of real and synthetic molecules, respectively. However, WGAN may not fully meet the requirements of clients who seek to generate synthetic molecules that not only retain similar chemical properties as real ones but also optimize user-defined metrics, such as drug-likeliness. This capability is crucial for facilitating the new drug discovery process. The primary limitation of WGAN in meeting this requirement is the absence of a mechanism to inform the generator about the performance of the generated molecules in the user-defined metrics. To address this limitation, we propose integrating a critic network into WGAN. The critic network concept is from RL to evaluate the state or state-action value. In our approach, the critic network is tasked with evaluating the metric values of generated molecules. Here, we focus on seven commonly used metric values output by the critic network given a molecule as input.

- **Validity** refers to the proportion of generated molecules that are valid, and the range is $[0, 100]$.
- **Uniqueness** refers to the percentage of valid molecules that exhibit unique characteristics, and the range is $[0, 100]$.
- **Novelty** refers to the proportion of valid molecules that differ from the ones in the existing dataset, and the range is $[0, 100]$.
- **Internal Diversity (IntDiv$_p$)** estimates the diversity of chemical compounds within the generated molecules, and is computed as:

$$\mathrm{IntDiv_p(G)} = 1 - \sqrt[p]{\frac{1}{|G|^2} \sum_{m_1, m_2 \in G} T(m_1, m_2)^p}. \quad (4)$$

  The range is $[0, 1]$.
- **Quantitative Estimation of Drug-likeliness (QED)** indicates the probability of a molecule being a potential candidate for a drug, and the range is $[0, 1]$.
- **Octanol-water partition coefficient (LogP)** refers to the relationship between a chemical's concentration in the octanol phase and that in the water phase within a two-phase octanol/water system, and the range is $[0, 1]$.
- **Similarity to a nearest neighbor (SNN)** calculates the average Tanimoto similarity $(m_G, m_R)$ between the fingerprints of a molecule $m_G$ from the generated set $G$ and its nearest neighbor molecule $m_R$ in the real dataset $R$. The range is $[0, 1]$.

### B. The cGraphGANFed Framework

We propose the cGraphGANFed framework, which integrates a critic network into our previously proposed GraphGANFed framework [14]. The goal is to ensure that synthetic molecules not only retain similar chemical properties as real ones but also optimize user-defined metrics. Fig. 1 shows the overview of cGraphGANFed, which comprises four major parts, i.e., a generator, discriminator, critic, and FL system.

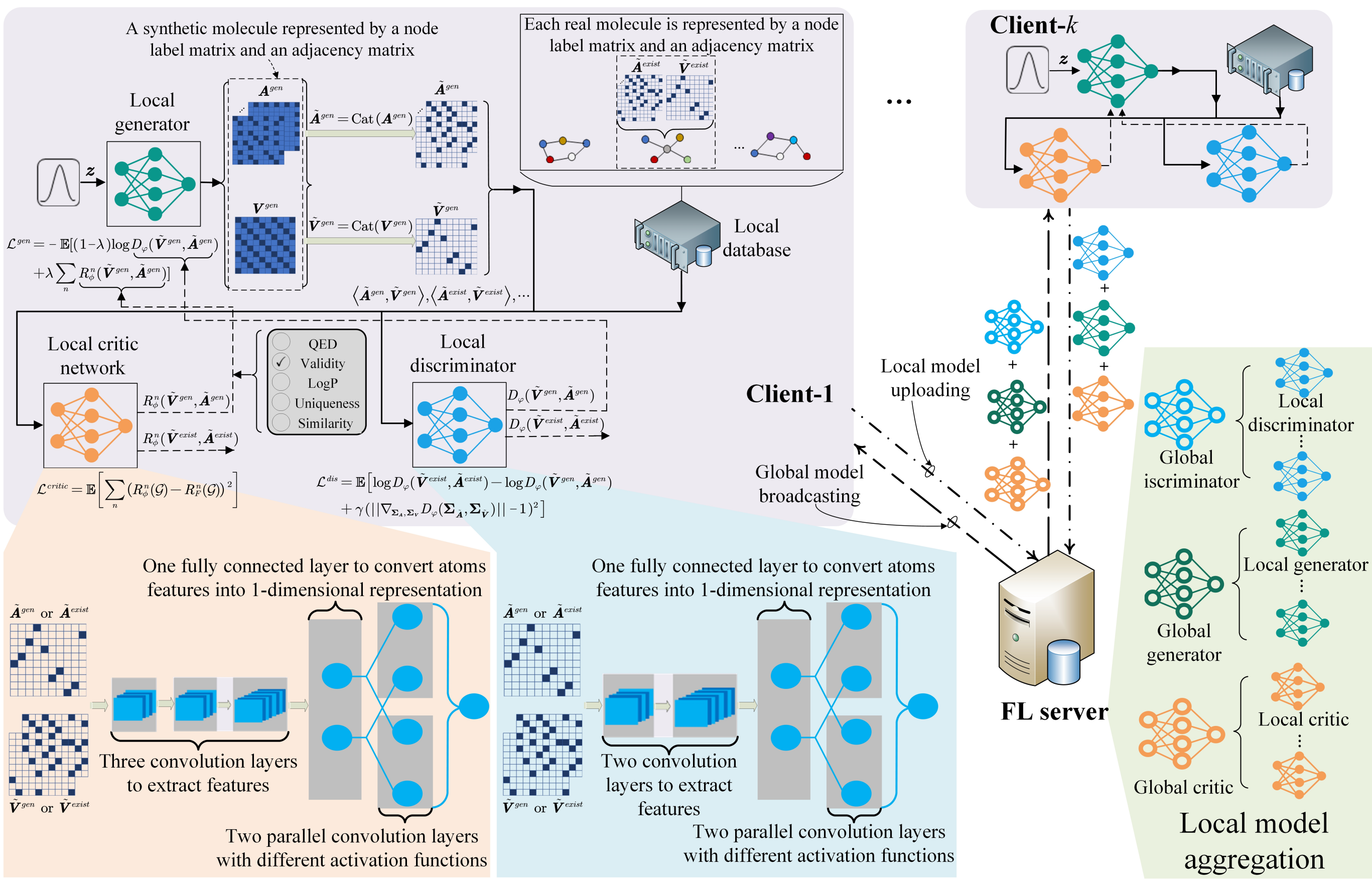

Fig. 1. The cGraphGANFed architecture.

*1) **Generator**:* The generator is implemented as a multi-layer perceptron (MLP). Its input is a random noise vector $\boldsymbol{z}$, sampled from a standard normal distribution. The generator outputs a molecular graph $\mathcal{G}$ represented by two continuous matrices: node label matrix $\boldsymbol{V}$ and adjacency matrix $\boldsymbol{A}$, with dimensions $\boldsymbol{V} \in \mathbb{R}^{N\times M}$ and $\boldsymbol{A} \in \mathbb{R}^{N\times N\times T}$, respectively. Here, $N$ represents the total number of atoms in a molecule, $M$ denotes the number of possible atom types, and $T$ is the bond types. Both $\boldsymbol{V}$ and $\boldsymbol{A}$ are probabilistically represented, with each node and edge type expressed through probabilities within categorical distributions across different types. To generate a molecule, discrete sparse samples represented by $\tilde{\boldsymbol{V}}$ and $\tilde{\boldsymbol{A}}$ are obtained through categorical sampling from $\boldsymbol{V}$ and $\boldsymbol{A}$, respectively. Due to the non-differentiability of the discrete samples, we used a direct gradient based on the categorical reparameterization using the Gumbel-Softmax [28] to enable gradient-based training. Specifically, samples from the categorical distribution are used during the forward pass, i.e., $\tilde{\boldsymbol{V}}_i = \mathrm{Cat}(\boldsymbol{V}_i)$ and $\tilde{\boldsymbol{A}}_{ij} = \mathrm{Cat}(\boldsymbol{A}_{ij})$, where $\tilde{\boldsymbol{V}}_i$ is the node label vector for node $i$ and $\tilde{\boldsymbol{A}}_{ij}$ is the bond between nodes $i$ and $j$ in the molecule, and original continuous matrices $\boldsymbol{V}$ and $\boldsymbol{A}$ are used during the backward pass. In our setting, the generated molecular graph $\mathcal{G}$ consists of a maximum of $N = 10$ atoms, $M = 10$ atom types (i.e., Carbon (C), Nitrogen (N), Oxygen (O), Fluorine (F), Bromine (Br), Phosphorus (P), Sulfur (S), Chlorine (Cl), Iodine (I), and one padding symbol (*)), and $T = 5$ bond types (i.e., zero, single, double, triple, and aromatic bonds). We apply the tanh activation function to each hidden layer in the MLP. The output of the final hidden layer is linearly projected to align with the dimensions of $\boldsymbol{X}$ and $\boldsymbol{A}$. The generator is trained to minimize the weighted combination of the Wasserstein loss [25] and the predicted user-defined metric values.

$$\mathcal{L}^{gen} = -\mathbb{E}[(1-\lambda)\log D_{\varphi}(\tilde{\boldsymbol{V}}^{gen}, \tilde{\boldsymbol{A}}^{gen}) + \lambda \sum_{n} R^{n}_{\phi}(\tilde{\boldsymbol{V}}^{gen}, \tilde{\boldsymbol{A}}^{gen})], \tag{5}$$

where $\tilde{\boldsymbol{V}}^{gen}$ and $\tilde{\boldsymbol{A}}^{gen}$ represent the node label and adjacency matrices of a generated molecule, respectively, $D_{\varphi}(\tilde{\boldsymbol{V}}^{gen}, \tilde{\boldsymbol{A}}^{gen})$ denotes the output of the discriminator for the generated molecule, $\lambda \in [0, 1]$ is the hyperparameter that regulates the trade-off between optimizing the generated molecules to retain similar chemical properties as real ones and maximizing the user-defined metric values, and $R^{n}_{\phi}(\tilde{\boldsymbol{V}}^{gen}, \tilde{\boldsymbol{A}}^{gen})$ is the normalized value of user-defined metric $n$ for the generated molecule represented by $\tilde{\boldsymbol{V}}^{gen}$ and $\tilde{\boldsymbol{A}}^{gen}$. Here, the value of user-defined metric $n$ is predicted by the critic network $R_{\phi}$ parameterized by $\phi$, as will be illustrated later. If, for instance, clients aim to generate synthetic molecules that optimize the Validity, QED, LogP, Similarity and Uniqueness metrics, then the loss function of the generator can be constructed as

$$\mathcal{L}^{gen} = -\mathbb{E}[(1-\lambda)\log D_{\varphi}(\tilde{\boldsymbol{V}}^{gen}, \tilde{\boldsymbol{A}}^{gen}) + \lambda(R_{\phi}^{valid} + R_{\phi}^{qed} + R_{\phi}^{logP} + R_{\phi}^{unique} - R_{\phi}^{sim})], \quad (6)$$

where $R_{\phi}^{valid}$, $R_{\phi}^{qed}$, $R_{\phi}^{logP}$, $R_{\phi}^{sim}$ and $R_{\phi}^{unique}$ are the predicted values for Validity, QED, LogP, Similarity and Uniqueness, respectively. Here, we employ a negative sign for Similarity because our objective is to minimize the similarities between the generated and real molecules.

*2) **Discriminator**:* The discriminator's objective is to distinguish between the generated and real molecules. We construct the discriminator using the GCN architecture outlined in [9]. Specifically, we employ the relational GCN (R-GCN) encoder, a convolutional network capable of processing non-directional and multi-edge graphs as inputs. The loss function of the discriminator is

$$\mathcal{L}^{dis} = \mathbb{E}\Big[\log D_{\varphi}(\tilde{\boldsymbol{V}}^{exist}, \tilde{\boldsymbol{A}}^{exist}) - \log D_{\varphi}(\tilde{\boldsymbol{V}}^{gen}, \tilde{\boldsymbol{A}}^{gen}) + \gamma(||\nabla_{\boldsymbol{\Sigma}_{\tilde{\boldsymbol{A}}}, \boldsymbol{\Sigma}_{\tilde{\boldsymbol{V}}}} D_{\varphi}(\boldsymbol{\Sigma}_{\tilde{\boldsymbol{A}}}, \boldsymbol{\Sigma}_{\tilde{\boldsymbol{V}}})|| - 1)^2\Big] \quad (7)$$

where $\tilde{\boldsymbol{V}}^{exist}$ and $\tilde{\boldsymbol{A}}^{exist}$ represent the node label and adjacency matrices of a real molecule in the existing dataset, respectively, $D_{\varphi}(\tilde{\boldsymbol{V}}^{exist}, \tilde{\boldsymbol{A}}^{exist})$ signifies the output of the discriminator for the existing molecule and $\gamma(||\nabla_{\boldsymbol{\Sigma}_{\tilde{\boldsymbol{A}}}, \boldsymbol{\Sigma}_{\tilde{\boldsymbol{V}}}} D_{\varphi}(\boldsymbol{\Sigma}_{\tilde{\boldsymbol{A}}}, \boldsymbol{\Sigma}_{\tilde{\boldsymbol{V}}})||-1)^2$ denotes the gradient penalty term, designed to stabilize the gradients of the discriminator. Here, $\gamma$ is the gradient penalty coefficient, $\boldsymbol{\Sigma}_{\tilde{\boldsymbol{A}}} = \varepsilon\tilde{\boldsymbol{A}}^{exist} + (1-\varepsilon)\tilde{\boldsymbol{A}}^{gen}$, and $\boldsymbol{\Sigma}_{\tilde{\boldsymbol{V}}} = \varepsilon\tilde{\boldsymbol{V}}^{exist} + (1-\varepsilon)\tilde{\boldsymbol{V}}^{gen}$, where $\varepsilon \in [0,1]$ is a predefined hyperparameter.

The discriminator architecture consists of two convolutional layers responsible for extracting atom features. Subsequently, a single-layer MLP computes a one-dimensional feature representation. This representation is simultaneously input into two parallel hidden layers. The final output combines the results of these parallel hidden layers through element-wise multiplication, followed by a tanh activation.

*3) **Critic network**:* The critic network is designed to predict the values of the metrics for the input molecule represented as a graph. The critic network, denoted as $R_{\phi}(\cdot)$ with parameter $\phi$, is also implemented based on the R-GCN encoder, similar to the discriminator. Specifically, it comprises three convolution layers, followed by one-layer MLP. The output of this MLP is then fed into two parallel MLP layers. The final output combines features from these parallel layers and passes them through a sigmoid activation function to generate different user-defined metric values. The critic network is trained to minimize the loss, which is defined as the mean squared error between the predicted and the ground truth metric values, i.e.,

$$\mathcal{L}^{critic} = \mathbb{E}\left[\sum_{n}\left(R_{\phi}^{n}(\mathcal{G}) - R_{F}^{n}(\mathcal{G})\right)^2\right] \quad (8)$$

where $R_{\phi}^{n}(\mathcal{G})$ and $R_{F}^{n}(\mathcal{G})$ are the predicted value output by the critic network and ground truth, respectively, for metric $n$ given an input molecular graph $\mathcal{G}$ represented by discrete node label and adjacency matrices, i.e., $\mathcal{G} = \left\langle \tilde{\boldsymbol{V}}^{gen}, \tilde{\boldsymbol{A}}^{gen} \right\rangle$ or $\left\langle \tilde{\boldsymbol{V}}^{exist}, \tilde{\boldsymbol{A}}^{exist} \right\rangle$. Note that these ground truth metric values are computed using various well-developed but computationally expensive metric functions [29, 30].

*4) **FL system**:* The FL system in cGraphGANFed comprises a centralized FL server and a group of clients. This system enables distributed training of the generator, discriminator, and critic models using the clients' local datasets, all without sharing their datasets, thus preserving data privacy. In traditional FL settings, where only one model needs to be trained, the FL server broadcasts a global model to all the clients. Upon receiving the global model, each client trains the global model based on its local dataset to derive its local model and then uploads the local model to the FL server, which aggregates the received local model to generate a new global model for the next global round. The global round continues until the global model converges or the number of global rounds exceeds a predefined threshold [31].

Different from traditional FL settings, cGraphGANFed involves three models interacting with each other. This raises the question: what is the optimal sequence for training these models among clients? Here, two possible sequences are provided as examples. 1) Both the critic and discriminator models are trained among clients until convergence, and then the generator training is initialized among clients. This ensures that the critic and discriminator can provide accurate guidance before training the generator. 2) Three models are trained simultaneously in each global round until all are converged. After extensive testing, we discovered that training all three models concurrently among clients in each global round yields the best performance. The reason is that a delicate balance among the performance of the generator, discriminator, and critic in cGraphGANFed is necessary to achieve stable and convergent training. If this balance is disrupted, such as by having a powerful discriminator that can easily distinguish synthetic molecules from real ones, the generator's improvement may stagnate. Therefore, we designed the FL system in cGraphGANFed to enable concurrent training of all three models by all the clients in each global round, which comprises the following four steps.

- **Global model broadcasting**: The FL server broadcasts the global models $\theta^{global}$, $\varphi^{global}$, and $\phi^{global}$ to all the clients.
- **Local model training**: After receiving the global generator and discriminator, each client first partitions its local dataset of real molecules into $B$ batches of equal size and performs training for both the local discriminator and generator over $E$ epochs. As shown in Fig. 2. in each epoch, a client undergoes the following steps: i) client $k$ generates a batch of synthetic molecules using its local generator $\theta_k^{local}$; ii) the local discriminator $\varphi_k^{local}$ and critic $\phi_k^{local}$ are trained based on the batch of synthetic molecules and the first batch of real molecules in $B$ through batched gradients descent. Notably, the node and adjacency matrices of both the synthetic (i.e., $\tilde{\boldsymbol{V}}^{gen}$ and $\tilde{\boldsymbol{A}}^{gen}$) and existing molecules (i.e., $\tilde{\boldsymbol{V}}^{exist}$ and $\tilde{\boldsymbol{A}}^{exist}$) are discrete and non-differentiable, and thus back-propagation cannot be directly applied. As we mentioned before, a gradient estimator [28] is employed to convert the discrete and non-differentiable matrices into continuous and differentiable matrices (i.e., $\boldsymbol{V}^{gen}$, $\boldsymbol{A}^{gen}$, $\boldsymbol{V}^{exist}$, and $\boldsymbol{A}^{exist}$) using the Gumbel-Softmax distribution, thus enabling back-propagation. Meanwhile, both the local discriminator $\varphi_k^{local}$ and local critic network $\phi_k^{local}$ inputs the synthetic and real molecules, and outputs the scores (that imply the quality or "realness") and metric values of input molecules, respectively. iii) The local generator $\theta_k^{local}$ is trained based on the discriminator and critic outputs from step ii). Steps i)-iii) are repeated to train the local generator $\theta_k^{local}$, discriminator $\varphi_k^{local}$, and critic $\phi_k^{local}$ until the last batch of existing molecules in $B$ has been used for training.

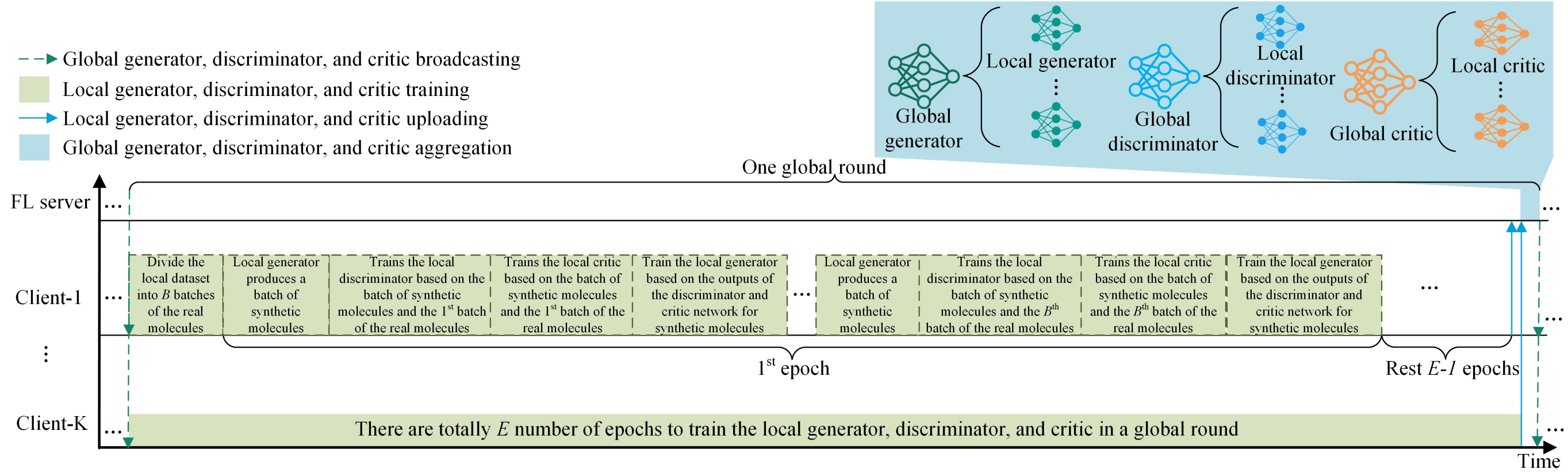

Fig. 2. Procedure of a global round in FL.

- **Local model Uploading:** After trained its local generator $\theta_k^{local}$, discriminator $\varphi_k^{local}$, and critic $\phi_k^{local}$ over $E$ epochs, client $k$ uploads $\theta_k^{local}$, $\varphi_k^{local}$ and $\phi_k^{local}$ to the FL server.
- **Local model aggregation:** After receiving $\theta_k^{local}$, $\varphi_k^{local}$ and $\phi_k^{local}$ from all clients, the FL server aggregates all the local generators, discriminators, and critic networks to update the global generator, discriminator, and critic network, respectively, based on, for example, FedAvg [31], i.e.,

$$\begin{cases} \theta^{global} = \sum_{k=1}^{K} \frac{\left|\boldsymbol{m}_k^{gen}\right|}{\left|\boldsymbol{M}^{gen}\right|} \theta_k^{local}, \\ \varphi^{global} = \sum_{k=1}^{k} \frac{\left|\boldsymbol{m}_k^{dis}\right|}{\left|\boldsymbol{M}^{dis}\right|} \varphi_k^{local}, \\ \phi^{global} = \sum_{k=1}^{k} \frac{\left|\boldsymbol{m}_k^{critic}\right|}{\left|\boldsymbol{M}^{critic}\right|} \phi_k^{local}, \end{cases} \tag{9}$$

where $\left|\boldsymbol{m}_k^{gen}\right|$, $\left|\boldsymbol{m}_k^{dis}\right|$ and $\left|\boldsymbol{m}_k^{critic}\right|$ represent the numbers of molecules used to train the local generator, discriminator and critic, respectively, for client $k$, and $\left|\boldsymbol{M}^{gen}\right| = \sum_{k=1}^{K} \left|\boldsymbol{M}_k^{gen}\right|$, $\left|\boldsymbol{M}^{dis}\right| = \sum_{k=1}^{K} \left|\boldsymbol{M}_k^{dis}\right|$ and $\left|\boldsymbol{M}^{critic}\right| = \sum_{k=1}^{K} \left|\boldsymbol{M}_k^{critic}\right|$.

## IV. SIMULATION SETUPS AND RESULTS

In this section, we extensively simulate to compare the performance of cGraphGANFed with GraphGANFed [14].

### A. Simulation setups

cGraphGANFed with GraphGANFed are both trained and tested based on three benchmark datasets, i.e., ESOL [32], QM8 [33], and QM9 [34]. Each dataset is randomly split into training, validation, and testing in a ratio of 80:10:10. The datasets are allocated to the clients based on independent and identically distribution (IID) and non-IID, and we will evaluate the performance of cGraphGANFed under these two distributions. Each molecule is labeled based on its molecular formula (e.g., $H_2O$, HF, etc.), and molecules with the same label are placed in the same class. In the IID scenario, molecules within each class are divided into $K$ groups of equal size, with each group assigned to a client. Conversely, in the non-IID setup, we randomly assign a varying number of molecules from each class to individual clients.

In all simulations, the generator is fed with a fixed-size vector $\boldsymbol{z}$ containing 16 elements, each sampled from a standard normal distribution, to produce a synthetic molecule. In each global ground, the number of local epochs $E$ to train a local model is the same for all the clients, i.e., $E = 5,000$ for ESOL and $E = 1,000$ for QM8 and QM9. The batch size is $B$ =16. The gradient penalty coefficient is $\gamma = 10$ to compute the discriminator's loss in Eq. (7). The generator, discriminator, and critic models are trained using the Adam optimizer with exponential decay rates $\beta_1 = 0.5$ and $\beta_2 = 0.999$. $\lambda$ in Eq. (5) is 0.5 and the learning rate is 0.0001 for the local generator, discriminator, and critic training. The source code of cGraphGANFed is available at https://github.com/danielmanu93/Conditional-GraphGANFed.

### B. Simulation results

*1) Critic convergence analysis:* Given the specified dimensions of the generator, discriminator, and critic models as ([32, 64, 128]), ([64, 128], 64, [128, 1]), and ([64, 64, 64], 64, [64, 1]), respectively, Fig. 3 shows 3 clients training these models over 100 global rounds based on QM8 for both IID and non-IID settings. The results show that all three models are converged in both IID and non-IID settings with similar convergence rates.

*2) Optimization of all the metrics:* Assume that the clients aim to optimize all the 7 metrics mentioned in Section III.A, and so the loss function of the generator includes the 7 metric values, i.e., $R_\phi^{valid}$, $R_\phi^{unique}$, $R_\phi^{nov}$, $R_\phi^{intDiv}$, $R_\phi^{qed}$, $R_\phi^{logP}$, and $R_\phi^{sim}$, output by the critic network. Note that in our previous GraphGANFed paper [14], we have discovered the optimal model structures of the generator and discriminator via numerous tests. These optimal structures may vary based on different dataset types and numbers of clients. The second and third columns in Tables I-V are the optimal model structures that we have discovered for the generator and discriminator, respectively, under different settings, and we will evaluate and compare the performance of cGraphGANFed and GraphGANFed based on these optimal model structures later on.

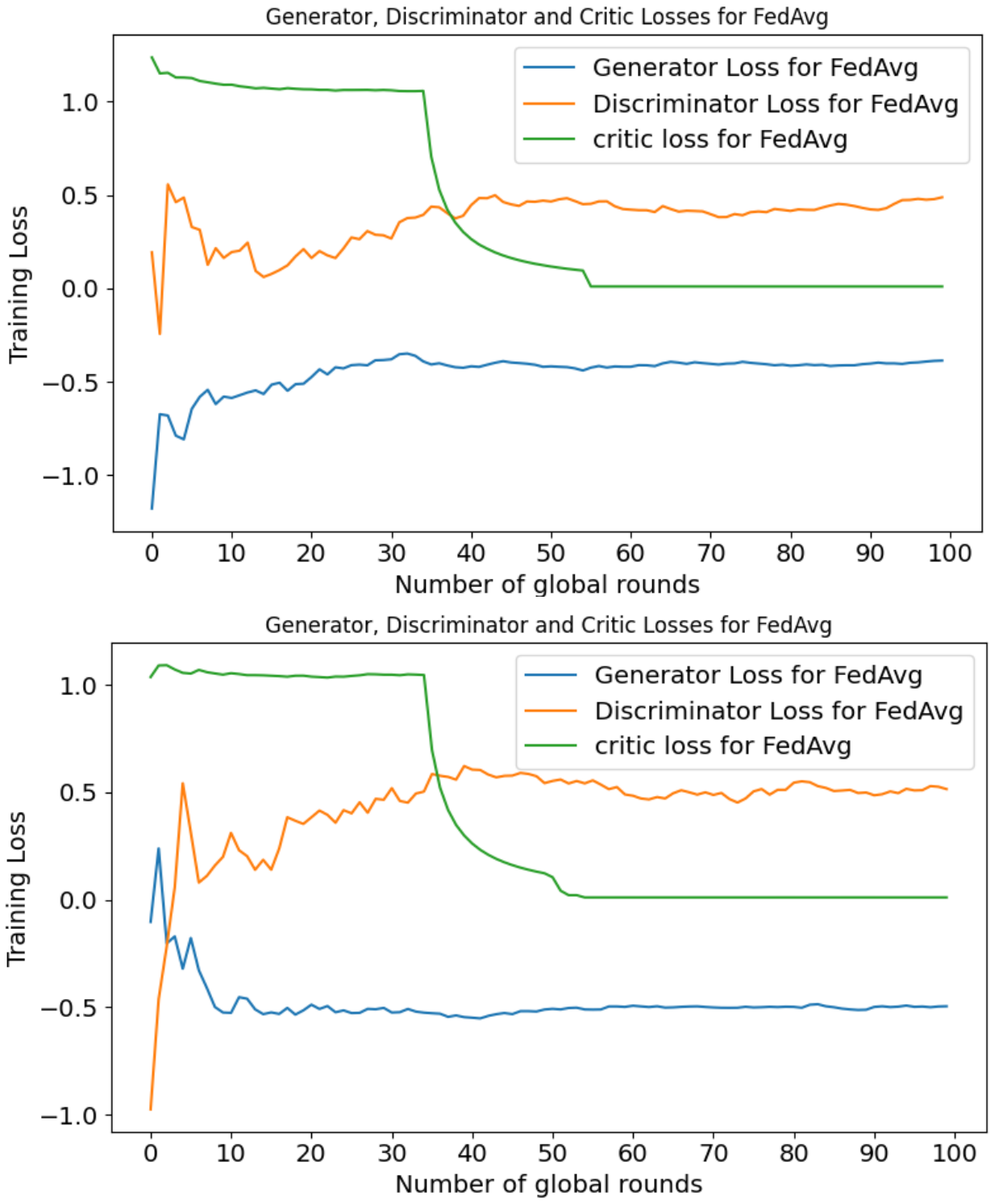

Fig. 3. Learning curves of the global generator, discriminator, and critic over the global rounds for IID (top) and non-IID (bottom).

Tables I-III show the performance comparison between cGraphGANFed and GraphGANFed [14] when the number of clients is 3, 4, and 5, respectively, in the IID setting. We can observe the following conclusions from the results. *First, cGraphGANFed significantly outperforms GraphGANFed in Validity and LogP, with a slight advantage in QED, across different numbers of clients and datasets.* For example, for the ESOL dataset, the Validity values for (cGraphGANFed vs GraphGANFed) are (**84.8** vs 73.2), (**83.9** vs 70.5), and (**73.2** vs 4.5) when the number of clients is 3, 4, and 5, respectively. The LogP values for (cGraphGANFed vs GraphGANFed) are (0.68 vs 0.68), (**0.76** vs 0.73), (**0.98** vs 0.88) when the number of clients is 3, 4, and 5. The QED values remain the same for GraphGANFed and cGraphGANFed. A similar conclusion can be drawn from the QM8 and QM9 datasets with a difference that cGraphGANFed has a slightly higher QED value than GraphGANFed when the number of clients is 5, i.e., (**0.48** vs 0.44), and (**0.5** vs 0.45) for QM8 and QM9, respectively, as shown in Table III. *Second, both cGraphGANFed and GraphGANFed produce synthetic molecules with consistently optimal Diversity (close to 1.0), Novelty (all equal to 100), and Similarity (close to 0) values across various client numbers and datasets.* This phenomenon can be attributed to the functionality of GraphGANFed, where a discriminator and generator play a min-max game, enabling the generator to generate synthetic molecules that can deceive the discriminator. This objective can be equivalent to optimizing Diversity, Novelty, and Similarity. *Third, the synthetic molecules generated by GraphGANFed have higher Uniqueness than cGraphGANFed.* Despite cGraphGANFed demonstrating lower performance in Uniqueness than GraphGANFed, this outcome can be attributed to an inverse correlation between Validity and Uniqueness. In other words, an increase in Validity inevitably results in a decrease in Uniqueness.

Tables IV-V show the performance comparison between cGraphGANFed and GraphGANFed [14] when the number of clients is 3 and 7, respectively, in the non-IID setting. The observations drawn here are the same as those derived from the IID scenario. An important observation emerges when comparing Validity and LogP values between IID and non-IID settings: both cGraphGANFed and GraphGANFed exhibit reductions. However, the decrease in Validity and LogP metrics for cGraphGANFed is notably smaller than that of GraphGANFed. This leads us to the conclusion that cGraphGANFed exhibits superior robustness to performance degradation in non-IID scenarios compared to GraphGANFed.

*3) Optimization of single metric:* In the previous simulations setup, where cGraphGANFed aims to optimize all the metrics, it was observed that cGraphGANFed only exhibited marginal enhancement in QED as compared to GraphGANFed. In this simulation, we refine the generator's loss function to exclusively target QED optimization, i.e., the loss function of the generator only includes $R_{\phi}^{qed}$.

Tables VI and VII show the performance of cGraphGANFed in the IID and non-IID settings, respectively. When we compare the performance of cGraphGANFed shown in Tables VI and VII with the performance of GraphGANFed shown in Tables I-V, we can conclude that *synthetic molecules generated by cGraphGANFed achieve consistently higher QED values than those generated by GraphGANFed, across different settings.* For instance, consider the QM9 dataset: GraphGANFed yields QED values of **0.50** for molecules generated under IID scenario with 3 and 4 clients, as depicted in the fifth row of Table I and Table II. Conversely, cGraphGANFed produces QED values of **0.55** and **0.56** for molecules generated under the same settings, as shown in the fifth and sixth rows of Table VI. This indicates a respective improvement of 10% and 12%, and we believe that further enhancement could be achieved by increasing the hyperparameter $\lambda$ in Eq. (5).

It is important to note that while cGraphGANFed achieves a 10% improvement in QED under the IID scenario with 3 clients, this enhancement may come with a trade-off in other metric values. For example, cGraphGANFed produces a lower LogP value, specifically 0.57 (as observed in the fifth row of Table VI), as compared to GraphGANFed, which yields 0.69 (also in the fifth row of Table VI), under the same settings.

*4) Resolution of Mode Collapse with Critic Network:* In GraphGANFed, we observe mode collapses if we use large discriminator models trained by the small dataset, i.e.,

ESOL. Here, mode collapses mean that synthetic molecules produced by the generator have very low Uniqueness (typically less than 1), i.e., the generator produces repetitive synthetic molecules. In cGraphGANFed, we observe that it is very robust to model collapse even if large discriminator models are applied. Tables VIII and IX present two instances where large discriminator models are employed under IID and non-IID settings, respectively. The findings from both tables indicate that cGraphGANFed is capable of preventing mode collapses by attaining 100 Uniqueness, in contrast to the 0.9 Uniqueness achieved by GraphGANFed, which implies the occurrence of mode collapses. Yet, owing to the inverse correlation between Validity and Uniqueness as we mentioned before, cGraphGANFed produces much lower Validity than GraphGANFed.

## V. Conclusions

In this paper, we have introduced cGraphGANFed, a novel extension to GraphGANFed designed to address the challenge of optimizing user-defined metrics in the generation of molecular structures while maintaining data privacy through FL. By integrating a critic network into the framework, cGraphGANFed empowers the generator to produce synthetic molecules that not only preserve similar chemical properties to real compounds but also excel in optimizing specific metrics crucial for drug discovery. Our extensive simulations showcase the efficacy of cGraphGANFed in two distinct scenarios. First, when tasked with optimizing all metrics simultaneously, cGraphGANFed exhibits superior performance over GraphGANFed, particularly excelling in Validity and LogP, with a slight improvement in QED. Second, when exclusively targeting the optimization of QED, cGraphGANFed produces synthetic molecules that outperform GraphGANFed by more than 10%, highlighting its ability to precisely target specific optimization objectives. Moreover, our results underscore the enhanced resilience of cGraphGANFed against challenges such as mode collapses and mitigation of performance reduction caused by non-IID. This resilience is pivotal for ensuring the reliability and robustness of the generated molecular structures, thereby bolstering their utility in the drug discovery pipeline.

TABLE I
PERFORMANCE COMPARISON BETWEEN cGRAPHGANFED AND GRAPHGANFED TOWARDS OPTIMIZATION OF ALL METRICS IN IID WHEN THE NUMBER OF CLIENTS IS 3.

| Algorithm | Dataset | Generator Dim. | Discriminator Dim. | Critic Dim. | QED | Diversity | Validity | Unique | Novelty | LogP | Sim |
|---|---|---|---|---|---|---|---|---|---|---|---|
| GraphGANFed | ESOL | [32,64] | [32,64],32,[64,1] | N/A | 0.52 | 1.00 | 73.2 | 98.8 | 100 | 0.68 | 0.0001 |
| cGraphGANFed | | | | [2,2,2],2,[2,1] | 0.52 | 1.00 | **84.8** | 95.8 | 100 | 0.68 | 0.0036 |
| GraphGANFed | QM8 | [64,128,256] | [64,128],64,[128,1] | N/A | 0.48 | 0.99 | 91.6 | 9.4 | 100 | 0.74 | 0.0059 |
| cGraphGANFed | | | | [64,64,64],64,[64,1] | 0.48 | 0.99 | **97.3** | 2.9 | 100 | **0.84** | 0.0005 |
| GraphGANFed | QM9 | [64,128,256] | [256,512],256,[512,1] | N/A | 0.5 | 0.99 | 72.2 | 10.9 | 100 | 0.69 | 0.0297 |
| cGraphGANFed | | | | [128,128,128],128,[128,1] | 0.5 | 0.99 | **89.1** | 4.7 | 100 | **0.71** | 0.0213 |

TABLE II
PERFORMANCE COMPARISON BETWEEN cGRAPHGANFED AND GRAPHGANFED TOWARDS OPTIMIZATION OF ALL METRICS IN IID WHEN THE NUMBER OF CLIENTS IS 4.

| Algorithm | Dataset | Generator Dim. | Discriminator Dim. | Critic Dim. | QED | Diversity | Validity | Unique | Novelty | LogP | Sim |
|---|---|---|---|---|---|---|---|---|---|---|---|
| GraphGANFed | ESOL | [32,128] | [32,64],32,[64,1] | N/A | 0.54 | 1.00 | 70.5 | 100 | 100 | 0.73 | 0.0146 |
| cGraphGANFed | | | | [2,2,2],2,[2,1] | 0.54 | 1.00 | **83.9** | 67.7 | 100 | **0.76** | 0.0308 |
| GraphGANFed | QM8 | [32,64,128] | [64,128],64,[128,1] | N/A | 0.49 | 1.00 | 78.4 | 18.1 | 100 | 0.54 | 0.0067 |
| cGraphGANFed | | | | [64,64,64],64,[64,1] | 0.49 | 1.00 | **81.2** | **18.6** | 100 | **0.70** | 0.0034 |
| GraphGANFed | QM9 | [64,128,256] | [256,512],256,[512,1] | N/A | 0.50 | 0.99 | 72.9 | 12.6 | 100 | 0.72 | 0.0056 |
| cGraphGANFed | | | | [128,128,128],128,[128,1] | 0.50 | **1.00** | **89.4** | 4.9 | 100 | 0.72 | 0.0008 |

TABLE III
PERFORMANCE COMPARISON BETWEEN cGRAPHGANFED AND GRAPHGANFED TOWARDS OPTIMIZATION OF ALL METRICS IN IID WHEN THE NUMBER OF CLIENTS IS 5.

| Algorithm | Dataset | Generator Dim. | Discriminator Dim. | Critic Dim. | QED | Diversity | Validity | Unique | Novelty | LogP | Sim |
|---|---|---|---|---|---|---|---|---|---|---|---|
| GraphGANFed | ESOL | [32,128] | [32,64],32,[64,1] | N/A | 0.43 | 1.00 | 4.5 | 100 | 100 | 0.88 | 0.0202 |
| cGraphGANFed | | | | [2,2,2],2,[2,1] | 0.43 | 1.00 | **73.2** | 100 | 100 | **0.98** | 0.0011 |
| GraphGANFed | QM8 | [32,64,128] | [64,128],64,[128,1] | N/A | 0.44 | 0.99 | 4.6 | 84 | 100 | 0.64 | 0.0281 |
| cGraphGANFed | | | | [64,64,64],64,[64,1] | **0.48** | 0.99 | **66.8** | 27.3 | 100 | **0.66** | 0.0000 |
| GraphGANFed | QM9 | [128,256,512] | [128,128],256,[128,1] | N/A | 0.45 | 1.00 | 87.9 | 11.9 | 100 | 0.34 | 0.0068 |
| cGraphGANFed | | | | [128,128,128],128,[128,1] | **0.5** | 1.00 | **89.00** | 4.2 | 100 | **0.70** | 0.0039 |

TABLE IV
PERFORMANCE COMPARISON BETWEEN cGRAPHGANFED AND GRAPHGANFED TOWARDS OPTIMIZATION OF ALL METRICS IN NON-IID WHEN THE NUMBER OF CLIENTS IS 3.

| Algorithm | Dataset | Generator Dim. | Discriminator Dim. | Critic Dim. | QED | Diversity | Validity | Unique | Novelty | LogP | Sim |
|---|---|---|---|---|---|---|---|---|---|---|---|
| GraphGANFed | ESOL | [32,128] | [32,64],32,[64,1] | N/A | 0.45 | 1.00 | 55.4 | 91.9 | 100 | 0.48 | 0.0015 |
| cGraphGANFed | | | | [2,2,2],2,[2,1] | 0.45 | 1.00 | **81.3** | 83.5 | 100 | 0.48 | 0.0003 |
| GraphGANFed | QM8 | [64,128,256] | [256,512],256,[512,1] | N/A | 0.49 | 0.99 | 88.9 | 10.4 | 100 | 0.62 | 0.0378 |
| cGraphGANFed | | | | [64,64,64],64,[64,1] | 0.49 | 0.99 | **93.1** | 7.7 | 100 | **0.7** | 0.0014 |
| GraphGANFed | QM9 | [64,128,256] | [256,512],256,[512,1] | N/A | 0.50 | 0.99 | 73.5 | 10.8 | 100 | 0.67 | 0.0078 |
| cGraphGANFed | | | | [128,128,128],128,[128,1] | 0.50 | **1.00** | **89.7** | 4.6 | 100 | **0.77** | 0.0032 |

TABLE V
PERFORMANCE COMPARISON BETWEEN cGRAPHGANFED AND GRAPHGANFED TOWARDS OPTIMIZATION OF ALL METRICS IN NON-IID WHEN THE NUMBER OF CLIENTS IS 7.

| Algorithm | Dataset | Generator Dim. | Discriminator Dim. | Critic Dim. | QED | Diversity | Validity | Unique | Novelty | LogP | Sim |
|---|---|---|---|---|---|---|---|---|---|---|---|
| GraphGANFed | ESOL | [32,128] | [32,64],32,[64,1] | N/A | 0.34 | 1.00 | 100.0 | 0.9 | 100 | 0.26 | 0.0011 |
| cGraphGANFed | | | | [2,2,2],2,[2,1] | **0.41** | 1.00 | 1.8 | **100.0** | 100 | **0.51** | 0.0381 |
| GraphGANFed | QM8 | [32,64,128] | [64,128],64,[128,1] | N/A | 0.44 | 0.99 | 20.9 | 59.3 | 100 | 0.65 | 0.0309 |
| cGraphGANFed | | | | [64,64,64],64,[64,1] | **0.46** | 0.99 | **49.1** | 35.9 | 100 | **0.66** | 0.0038 |
| GraphGANFed | QM9 | [128,256,512] | [128,128],256,[128,1] | N/A | 0.50 | 0.99 | 77.3 | 7.1 | 100 | 0.42 | 0.0039 |
| cGraphGANFed | | | | [128,128,128],128,[128,1] | **0.51** | **1.00** | **98.9** | 1.3 | 100 | **0.78** | 0.0000 |

TABLE VI
PERFORMANCE OF cGRAPHGANFED AND GRAPHGANFED TOWARDS OPTIMIZATION OF QED IN IID, WHERE CRITIC DIM. IN cGRAPHGANFED IS THE SAME AS THAT IN TABLE I/II FOR DIFFERENT DATASETS.

| Algorithm | Dataset | Generator Dim. | Discriminator Dim. | Clients | QED | Diversity | Validity | Unique | Novelty | LogP | Sim |
|---|---|---|---|---|---|---|---|---|---|---|---|
| cGraphGANFed | ESOL | [32,64] | [32,64],32,[64,1] | 3 | **0.58** | 1.00 | 76.8 | 95.4 | 100 | 0.45 | 0.0020 |
| GraphGANFed | | [32,128] | | 4 | **0.55** | 1.00 | **90.2** | **94.1** | 100 | **0.83** | 0.0008 |
| cGraphGANFed | QM8 | [64,128,256] | [64,128],64,[128,1] | 3 | **0.55** | 0.99 | 96.7 | 5.4 | 100 | 0.73 | 0.0064 |
| GraphGANFed | | [32,64,128] | | 4 | **0.54** | 0.99 | 79.3 | 17.6 | 100 | 0.57 | 0.0060 |
| cGraphGANFed | QM9 | [64,128,256] | [256,512],256,[512,1] | 3 | **0.55** | 0.99 | 61.0 | **17.5** | 100 | 0.57 | 0.0000 |
| GraphGANFed | | | | 4 | **0.56** | 1.00 | 96.4 | 2.7 | 100 | 0.72 | 0.0040 |

TABLE VII
PERFORMANCE OF CGRAPHGANFED AND GRAPHGANFED TOWARDS OPTIMIZATION OF QED IN IID, WHERE CRITIC DIM. IN CGRAPHGANFED IS THE SAME AS THAT IN TABLE IV/V FOR DIFFERENT DATASETS.

| Algorithm | Dataset | Generator Dim. | Discriminator Dim. | Clients | QED | Diversity | Validity | Unique | Novelty | LogP | Sim |
|---|---|---|---|---|---|---|---|---|---|---|---|
| cGraphGANFed | ESOL | [32,128] | [32,64],32,[64,1] | 3 | **0.55** | 1.00 | 44.6 | **100.0** | 100 | **0.79** | 0.0039 |
| GraphGANFed | | | | 7 | **0.56** | 1.00 | 0.9 | 100.0 | 100 | **0.73** | 0.0166 |
| cGraphGANFed | QM8 | [64,128,256] | [256,512],256,[512,1] | 3 | **0.59** | 0.99 | 84.1 | **15.5** | 100 | 0.64 | 0.0292 |
| GraphGANFed | | [32,64,128] | [64,128],64,[128,1] | 7 | **0.55** | 1.00 | **84.0** | 20.1 | 100 | **0.71** | 0.0042 |
| cGraphGANFed | QM9 | [64,128,256] | [256,512],256,[512,1] | 3 | **0.56** | 0.99 | 82.1 | **5.5** | 100 | **0.78** | 0.0201 |
| GraphGANFed | | [128,256,512] | [128,128],256,[128,1] | 7 | **0.55** | 1.00 | 82.9 | 7.7 | 100 | 0.57 | 0.0019 |

TABLE VIII
PERFORMANCE OF CGRAPHGANFED AND GRAPHGANFED BY APPLYING LARGE DISCRIMINATOR FOR ESOL IN IID.

| Algorithm | Generator Dim. | Discriminator Dim. | Critic Dim. | Clients | QED | Diversity | Validity | Unique | Novelty | LogP | Sim |
|---|---|---|---|---|---|---|---|---|---|---|---|
| GraphGANFed | [32,128] | [32,128],32,[128,1] | N/A | 4 | 0.34 | 1.00 | 100 | 0.9 | 100 | 0.26 | 0.0004 |
| cGraphGANFed | | [32,128],32,[128,1] | [2,2,2],2,[2,1] | | **0.37** | 1.00 | 2.7 | **100.0** | 100 | **0.85** | 0.0042 |
| GraphGANFed | [32,128] | [32,256],32,[256,1] | N/A | 4 | 0.34 | 1.00 | 100 | 0.9 | 100 | 0.26 | 0.0004 |
| cGraphGANFed | | [32,256],32,[256,1] | [2,2,2],2,[2,1] | | 0.34 | 1.00 | 0.9 | 100.0 | **100** | 0.26 | 0.0000 |

TABLE IX
PERFORMANCE OF CGRAPHGANFED AND GRAPHGANFED BY APPLYING LARGE DISCRIMINATOR FOR ESOL IN NON-IID.

| Algorithm | Generator Dim. | Discriminator Dim. | Critic Dim. | Clients | QED | Diversity | Validity | Unique | Novelty | LogP | Sim |
|---|---|---|---|---|---|---|---|---|---|---|---|
| GraphGANFed | [32,128] | [32,128],32,[128,1] | N/A | 3 | 0.34 | 1.00 | 100 | 0.9 | 100 | 0.26 | 0.0004 |
| cGraphGANFed | | [32,128],32,[128,1] | [2,2,2],2,[2,1] | | **0.38** | 1.00 | 11.6 | **100.0** | 100 | **0.92** | 0.001 |
| GraphGANFed | [32,128] | [32,256],32,[256,1] | N/A | 3 | 0.34 | 1.00 | 100 | 0.9 | 100 | 0.26 | 0.0004 |
| cGraphGANFed | | [32,256],32,[256,1] | [2,2,2],2,[2,1] | | 0.34 | 1.00 | 0.9 | **100.0** | 100 | 0.26 | 0.0000 |